\documentclass[preprint,review,12pt]{elsarticle}

\usepackage{amssymb}
\usepackage{array}
\usepackage{amsmath}
\usepackage{booktabs}
\usepackage{multirow}
\usepackage{colortbl}
\usepackage{xcolor} 
\usepackage{hyperref} 
\hypersetup{
	colorlinks=true,
	linkcolor=blue,
	citecolor=green,
	urlcolor=cyan
}

\usepackage{tabularx}

\begin{document}

\begin{frontmatter}



\title{Change-Aware Siamese Network for Surface Defects Segmentation under Complex Background}

\author[inst1]{Biyuan Liu} 
\author[inst1]{Huaixin Chen*}
\author[inst3]{Huiyao Zhan}
\author[inst1]{Sijie Luo}
\author[inst2]{Zhou Huang}
\author[inst4]{Hao Cheng}

\affiliation[inst1]{organization={School of Resources and Environment},
            addressline={University of Electronic Science and Technology of China}, 
            city={Chengdu},
            state={Sichuan},
            country={China}}

\affiliation[inst2]{
    organization={Sichuan Changhong Electric Co., Ltd.},
    city={Chengdu},
    state={Sichuan},
    country={China}}


\affiliation[inst3]{
    organization={South China Normal University},
    city={Shanwei},
    state={Guangdong},
    country={China}}
\affiliation[inst4]{
organization={University of Twente},
addressline={Hallenweg 8}, 
city={Enschede},
country={The Netherlands}}

\begin{abstract}
Despite the eye-catching breakthroughs achieved by deep visual networks in detecting region-level surface defects, the challenge of high-quality pixel-wise defect detection remains due to diverse defect appearances and data scarcity. To avoid over-reliance on defect appearance and achieve accurate defect segmentation, we proposed a change-aware Siamese network that solves the defect segmentation in a change detection framework. A novel multi-class balanced contrastive loss is introduced to guide the Transformer-based encoder, which enables encoding diverse categories of defects as the unified class-agnostic difference between defect and defect-free images. The difference presented by a distance map is then skip-connected to the change-aware decoder to assist in the location of both inter-class and out-of-class pixel-wise defects. In addition, we proposed a synthetic dataset with multi-class liquid crystal display (LCD) defects under a complex and disjointed background context, to demonstrate the advantages of change-based modeling over appearance-based modeling for defect segmentation. In our proposed dataset and two public datasets, our model achieves superior performances than the leading semantic segmentation methods, while maintaining a relatively small model size. Moreover, our model achieves a new state-of-the-art performance compared to the semi-supervised approaches in various supervision settings.
\end{abstract}





\end{frontmatter}


\section{Introduction}\label{sec1}

 Surface defect inspection is a crucial step in manufacturing applications to prevent potential quality issues, economic loss, and even safety problems. These defects can manifest in various forms, such as dirt, spots, and fractures. They are commonly found in a range of industrial products, encompassing steel \cite{tabernik2020segmentation, Huang_Wu_Xie_2021}, LED \cite{lin2019automated}, and magnetic tile \cite{liu2022semi}. Unlike semantic objects, the surface defects generally do not have a regular shape, clear interpretation, or continuous context with the background, which causes difficulties for empirically designed methods \cite{sime2023uncertainty}. To facilitate the automation of defect inspection, deep learning-based approaches have been applied in multi-level defect detection. (1) {\bf Image-level classification} in earlier works resort to classifying whether an image contains defects or not, without giving a specific pixel-wise location \cite{masci2012steel,faghih2016deep,racki2018compact}. In SegNet \cite{tabernik2020segmentation} and its variants \cite{bovzivc2021end,bovzivc2021mixed}, pixel-level annotations are introduced as auxiliary information to the network yet ultimately output the binary classification results. 
 (2) {\bf Defect localization at fuzzy level} refers to obtaining a relatively fine-grained output without pixel-wise supervision. For instance, the class activation map \cite{zhou2016learning} is utilized for locating the blurry LED defects \cite{lin2019automated} and industrial anomalies \cite{lin2021cam} with image-level supervision. The methods based on non-defective sample modeling  \cite{roth2022towards,hyun2024reconpatch,rudolph2021same}, focus on modeling the distribution of defect-free data in the training phase, and subsequently assess the deviations in the distribution between anomaly and normal samples. The reconstruction-based anomaly detection approaches \cite{bergmann2019mvtec,schlegl2017unsupervised,batzner2024efficientad} aim at precisely reconstructing instances of normal data. The anomalies are figured out by noting these regions where the model fails to accurately reconstruct them \cite{ruff2021unifying}. While these methods do not necessitate a substantial volume of training data, the absence of meticulous supervision results in imprecise pixel-level predictions.
 (3) {\bf Fine-grained segmentation} has been increasingly applied for defect detection \cite{gao2019faster,he2019fully,Du_Shen_Fu_2021,dong2019pga, Huang_Wu_Xie_2021}. there exists a paradox between striving for zero defect manufacturing \cite{caiazzo2022towards} and the availability of sufficient defective samples.   
 To alleviate the shortage of pixel-label annotations, various studies have introduced additional priors, including visual saliency \cite{luo2023maminet}, repeat pattern analysis \cite{huang2022rpdnet}, and interactive click \cite{Du_Shen_Fu_2021}. Additionally, these studies have embraced semi-supervised techniques such as pseudo labeling \cite{sime2023uncertainty,xu2022efficient} and consistency regularization \cite{sime2022semi}, to further enhance their approaches.

 \begin{figure}[!t]%
 	\centering
\includegraphics[width=1.0\textwidth]{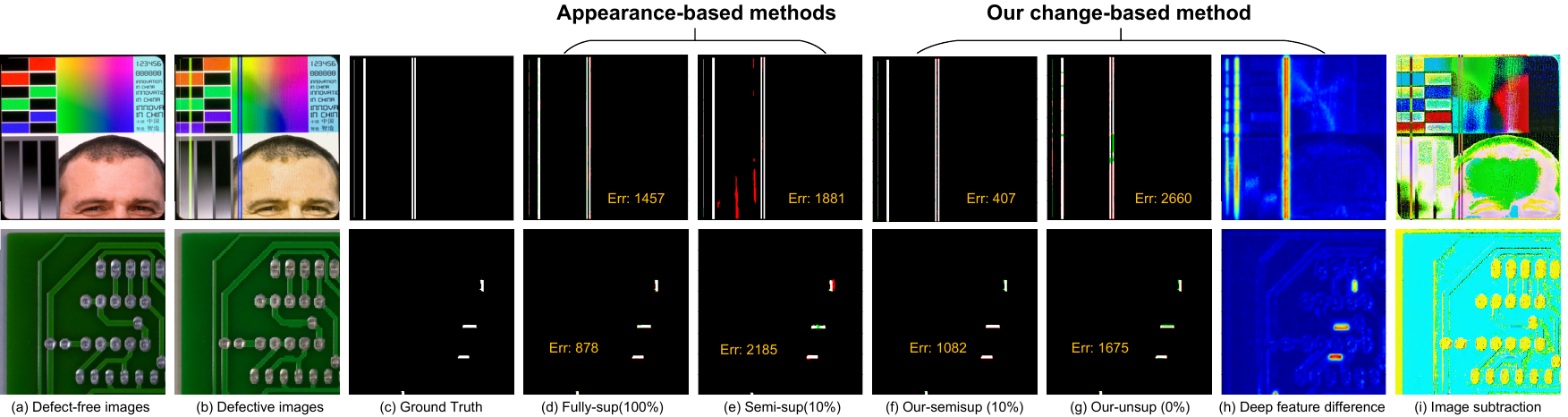}
 	\caption{The examples illustrate how our change-based and appearance-based methods have segmented defects in fully-supervised, semi-supervised, and unsupervised settings. The results in column (d) are derived from SegFormer \cite{SegFormer}. The outcomes in column (e) originate from UAPS \cite{sime2023uncertainty}. In the prediction maps, {\color{green}green} signifies missed detections and {\color{red}red} indicates erroneous detections. The term "Err" quantifies the total of these errors. Our model outperforms semi-supervised methods and achieves competitive outcomes using only 10\% of the training samples compared to the fully-supervised model. }\label{first_figure}
 \end{figure}

However, these aforementioned methods that locate defects based on appearance priors are not reliable due to the inherent contradiction between data scarcity and diverse manifestations of defects (see Figure \ref{first_figure}). Limited defect samples can yield a skewed representation of the true data distribution, subsequently leading to deteriorated generalization performance in these appearance-based methods \cite{sime2023uncertainty}.
It should be emphasized that locating defects based on their visual characteristics in products, such as printed circuit boards (PCBs), liquid crystal displays (LCDs), and printed publications, constitutes a substantial challenge. The complex and occasionally ambiguous patterns of the background can obscure these defects, consequently increasing the complexity of their detection.

Our motivation to transform defect detection as a change detection problem is based on two self-evident facts: (1) Obtaining defect-free samples is considerably easier than acquiring defect images. (2) Defect regions essentially correspond to the differences between defect-free and defective samples. Identifying defective regions proves challenging without a clean reference even for human observers. In this regard, we propose an accurate defect segmentation method based on data simulation and change feature modeling. This approach is particularly effective for surface defects with relatively steady but complex background patterns, such as PCB, LCD, and printed publications. 
 
More specifically, we propose a novel change-aware Siamese network with a change attention mechanism to solve pixel-wise defect detection. 
In the encoding stage, a Transformer-based Siamese network constrained by multi-class balanced contrastive loss (BCL) is employed to extract the difference features between the clean and the defective samples. Then, the hierarchical Siamese feature pairs are fused by multi-stage subtraction and upsampled to a high resolution. In the decoding stage, the feature distance map is skipped-connected to the decoder and acts as a change region attention to assist in locating the pixel-wise defects.
This change attention mechanism is applied using addition for intra-class detection and multiplication for our-of-class (OOC) detection.
Opposed to directly modeling the defect appearance, our proposed method models the defects as differences between defect-free and defective images, which empowers the generalization of detecting unseen defects. 

Furthermore, the community dedicated to surface defect detection requires a challenging dataset. The predominance of smaller datasets obstructs the thorough evaluation of current models. For instance, the average precision for commonly utilized datasets such as KolektorSSD \cite{bovzivc2021mixed}, DAGM2007 \cite{bovzivc2021end}, and Severstal-Steel \cite{bovzivc2021mixed} has attained the levels of 100$\%$, 100$\%$, and 98.7$\%$, respectively. Given the rapid ascension of LCDs as a leading display technology with extensive use in computers and mobile phones, we introduce a novel dataset aimed at enhancing LCD defect detection.

To summarize, our contributions are as follows:
\begin{itemize}
	\setlength{\itemsep}{0pt}
	\setlength{\parsep}{0pt}
	\setlength{\parskip}{0pt}
	\item We propose a change-aware Siamese network for defect segmentation. The modeling mechanism relies on changing features between clean and defective images instead of defect appearance, providing the possibility for synthetic data supervision and unseen class generalization. 
	\item In the encoding stage, the Transformer-based encoder supervised by balanced contrastive loss learns multi-class balanced feature differences between defective and defect-free images. In the decoding stage, the change-aware decoder leverages the feature discrepancies for enhanced accuracy and robustness in defect localization. 
	\item To facilitate the training and evaluation of our change-aware model, we introduce a synthetic LCD defect dataset named SynLCD. It serves as a benchmark to compare our model against other segmentation methods. 
    \item The experiments in SynLCD, PKU-Market-PCB \cite{ding2019tdd}, and MvTec-AD \cite{bergmann2019mvtec} datasets demonstrate that our network surpasses the state-of-the-art (SOTA) appearance-based segmentation methods. Furthermore, the comparison involving five SOTA semi-supervised segmentation methods highlights our model's superiority across different supervision levels. 
\end{itemize}

The remaining sections of this paper are organized as follows: Section \ref{sec2} presents related work about defect detection and change detection methods. We formulate our change-modeling network in section \ref{sec3} and conduct an extensive comparison with state-of-the-art fully-supervised and semi-supervised defect segmentation models in terms of intra-class and out-of-class performance in section \ref{sec4}. Finally, Section \ref{sec5} concludes this paper.


\section{Related Works}\label{sec2}
In this section, we introduce surface defect detection at various levels of detection granularity, along with change detection methods. The work most relevant to our study involves reconstruction-and-differencing based anomaly detection methods. These methods identify the approximate location of general surface defects with a differencing process between reconstructed and input images. In contrast, we employ deep feature change detection instead of simple differencing in the image space. Our focus is on precise segmentation in scenarios where defects can be subtle and potentially obscured during the reconstruction process. This focus is crucial to maintaining our primary emphasis on the core issue.


\subsection{Surface Defect Detection}\label{subsec2-1}
\textbf{Image-label detection.} 
Masci et al. \cite{masci2012steel} applied CNN to steel surface defect detection, highlighting CNN's superiority over manual features. Faghih-Roohi et al. \cite{faghih2016deep} explored the impact of network complexity on defect detection performance. Racki et al. \cite{racki2018compact} introduced a compact CNN for detecting synthetic textured anomalies by incorporating auxiliary segmentation labels alongside the classification task. SegNet \cite{tabernik2020segmentation} refined this approach by merging the distinct stages of segmentation and classification into an end-to-end training framework. Bo{\v{z}}i{\v{c}} et al. \cite{bovzivc2021mixed} embarked on an exploration of the impact of varying levels of supervision, from weak to full, on the accuracy of defect classification. Despite these advancements, early deep learning-based research primarily focused on image-level defect detection, with limited attention to pixel-wise defect localization.

\textbf{Fuzzy and region level detection.} Limited by the pixel-wise annotations in the anomaly detection task, some studies seek to consult the weak-supervised \cite{lin2019automated,lin2021cam} and unsupervised learning \cite{ruff2021unifying}. Class activation map \cite{zhou2016learning} is widely used to indicate the potential anomalous regions among an image with only image-level hints \cite{lin2019automated,lin2021cam}. However, this merely eases annotation labor but fails to address the fundamental issue of data scarcity. On the other hand, the wealth of defect-free data greatly prompts the advancement of non-defective modeling and reconstruction-based methods. {The \textit{non-defective modeling}} focuses on building an embedding model of normal samples and identifying the anomaly instances by measuring their deviation from the latent space. Defects are fuzzily spotted by patch-wise representation (e.g., PatchCore \cite{roth2022towards} and ReconPatch \cite{hyun2024reconpatch}), receptive field upsampling \cite{deng2022anomaly}, and gradient back-propagation in normalizing-flow based model \cite{rudolph2021same}.
{The \textit{reconstruction-based model}} is typically trained to reconstruct defect-free samples and identify anomalies, while it fails to generate the instances. The autoencoder \cite{bergmann2019mvtec} and generative-adversarial network (GAN) \cite{schlegl2017unsupervised} are commonly employed in the reconstruction process. A straightforward differencing process between the input and reconstructed samples is applied for obtaining defect region, such as the element-wise square distance in EfficientAD \cite{batzner2024efficientad}. However, a common issue is the occurrence of false-positive detections triggered by imprecise reconstructions of normal images. To sum up, due to the absence of pixel-wise annotations for these methods, it remains unclear which image points are anomalies, leading to indistinct detection results. 

\textbf{Pixel-wise detection.} Recently, there has been a growing focus on pixel-level defect detection extended of semantic segmentation models. He et al. \cite{he2019fully} proposed to locate wood defects by adopting the FCN architecture \cite{long2015fully}. Huang and Xiang \cite{huang2022rpdnet} adapted the DeepLab v3+ architecture \cite{deeplabv3plus2018} with minor modifications for the fabric defect segmentation. Du et al. \cite{du2020automatic} extended the U-Net \cite{2015unet} into a two-stream structure for segmenting defects in X-ray images. More recently, attention mechanisms have been employed for modeling local and global contextual dependencies. Dong et al. \cite{dong2019pga} proposed to segment steel surface defects with global context attention. Yeung et al. \cite{yeung2023attentive} refined SegFormer \cite{xie2021segformer} with a boundary-aware module for Transformer-based defect segmentation.
Defect segmentation enhances understanding of defective samples but is constrained by the cost of fine-grained labels.
   
Therefore, some recent studies resort to semi-supervised techniques such as pseudo labeling \cite{sime2023uncertainty,xu2022efficient} and consistency regularization \cite{sime2022semi}. 
Pseudo-labeling methods \cite{chen2021semi,fan2022ucc} generate pseudo-labels for unlabeled samples via a pretrained network, potentially enhancing model performance with these additional training signals. However, the predictive noise in unlabeled samples can compromise pseudo-label quality, thereby constraining their utility.
Consistency regularization posits that model predictions for unlabeled samples should remain consistent under controlled perturbations, aiming to minimize prediction discrepancies in different scenarios. Various heuristics have been introduced for consistency regularization, such as co-training \cite{qiao2018deep}, mean teacher \cite{yu2019uncertainty}, and multi-head prediction uncertainty \cite{sime2023uncertainty}.
We provide a comparison between these semi-supervised methods and our change-modeling architecture given limited labeled samples in Table \ref{tab:my_label}.

\subsection{Change Detection}\label{sec_changedetection}

Image change detection is designed to identify pre-defined differences between the images captured at different times \cite{liu2022lsnet}. The primary challenge in change detection lies in differentiating semantic changes from noisy alterations, including variations in illumination, saturation changes, and disturbances from irrelevant backgrounds. \cite{liu2023transformer}. It is widely applied in handwritten signature verification \cite{bromley1993signature}, street scene \cite{guo2018learning}, and remote sensing change detection \cite{liu2022lsnet}. 
In ChangChip \cite{fridman2021changechip}, surface defects in PCB are identified through manual image registration and comparison. However, it entails prolonged preprocessing times and necessitates hyperparameter fine-tuning for image subtraction. Zagoruyko et al. \cite{zagoruyko2015learning} pioneered the application of CNN for image comparison. Daudt et al. \cite{RC2019Fully} further developed an FCN-based Siamese architecture to enable arbitrary-sized image change detection. Several studies \cite{CD2017contra, guo2018learning} have concentrated on introducing contrastive loss \cite{2006contra}, a pivotal aspect for minimizing the distance of unchanged feature pairs while maximizing the distance of changed feature pairs. However, these contrastive approaches are primarily designed for binary changes and cause imbalance attention for different change categories, as illustrated in Figure \ref{fig_ablation}.

In our research context, the most relevant studies are background reconstruction methods \cite{sae2022semi,lv2020novel}. These work innovatively reconstructs flawless images from unlabeled data and employs a differential mapping technique between the original and reconstructed images to obtain the final segmentation map. However, the quality of the reconstructed image and image-level differencing become their bottlenecks.

\section{Method}\label{sec3}
\subsection{Problem definition: Appearance-modeling vs. Change-modeling}\label{sec-appvschange}

Industrial materials like LCD, PCB, and printed products (e.g. books, drawings, and trademarks), exhibit relatively consistent appearances and surface patterns when they are defect-free. Based on this observation, we simplify the formation process of surface defect images, represented as $x_{ng}$ (where "ng" stands for "not good"). This involves overlaying a standard clean image $x_\text{ok}$, with $x_\text{defect}$ in a specific manner, followed by a global nonlinear transformation.
This process can be formulated as 
\begin{equation} 
	x_\text{ng} = \sigma(x_\text{ok}\oplus {x_\text{defect}})\label{eq1},
\end{equation}
where $\sigma$ represents a nonlinear global transformation (e.g. material batch differences, aging, lighting, and imaging distortion), $\oplus$ indicates some kind of overlaying way (e.g. corrosion, breakage, mixing, and direct covering). For the classical segmentation paradigms, the model $f'$ identifies defect objects based on their appearance and context, which can be formulated according to the assumption of equation \eqref{eq1} as
\begin{equation} \label{eq2}
	\hat{x}_\text{defect} = f'(x_\text{ng}) =f'(\sigma(x_\text{ok}\oplus {x_\text{defect}})).
\end{equation}
It implies that the model $f'$ is required to separate $\hat{x}_\text{defect}$ from complex background $x_\text{ok}$ under nonlinear interference $\sigma$. However, the background content may closely resemble defects, as depicted in Figure \ref{fig2} (g), rendering the distinction based on defect appearance unreliable. We aim to model the defect in defective images as difference from defect-free ones, which is
\begin{equation}
	\begin{split}
		\hat{x}_\text{defect}&=f(x_\text{ng},\hat{x_\text{ok}}), \\ 
		& =\sigma({{x}_\text{ok}}\oplus {{x}_\text{defect}})\ominus {\hat{{x}_\text{ok}}}.  
	\end{split}
\end{equation}
In the change-modeling paradigm, the model learns a deep subtraction function $\ominus$, overcoming limitations associated with defect appearance. The disturbance of the nonlinear transformation $\sigma$ and complex background is mitigated with the aid of the easily obtainable defect-free image $\hat{x}_\text{ok}$.

\subsection{Change-aware Siamese Network}\label{subsec3-2}

\begin{figure}[h]%
	\centering
	\includegraphics[width=1.0\textwidth]{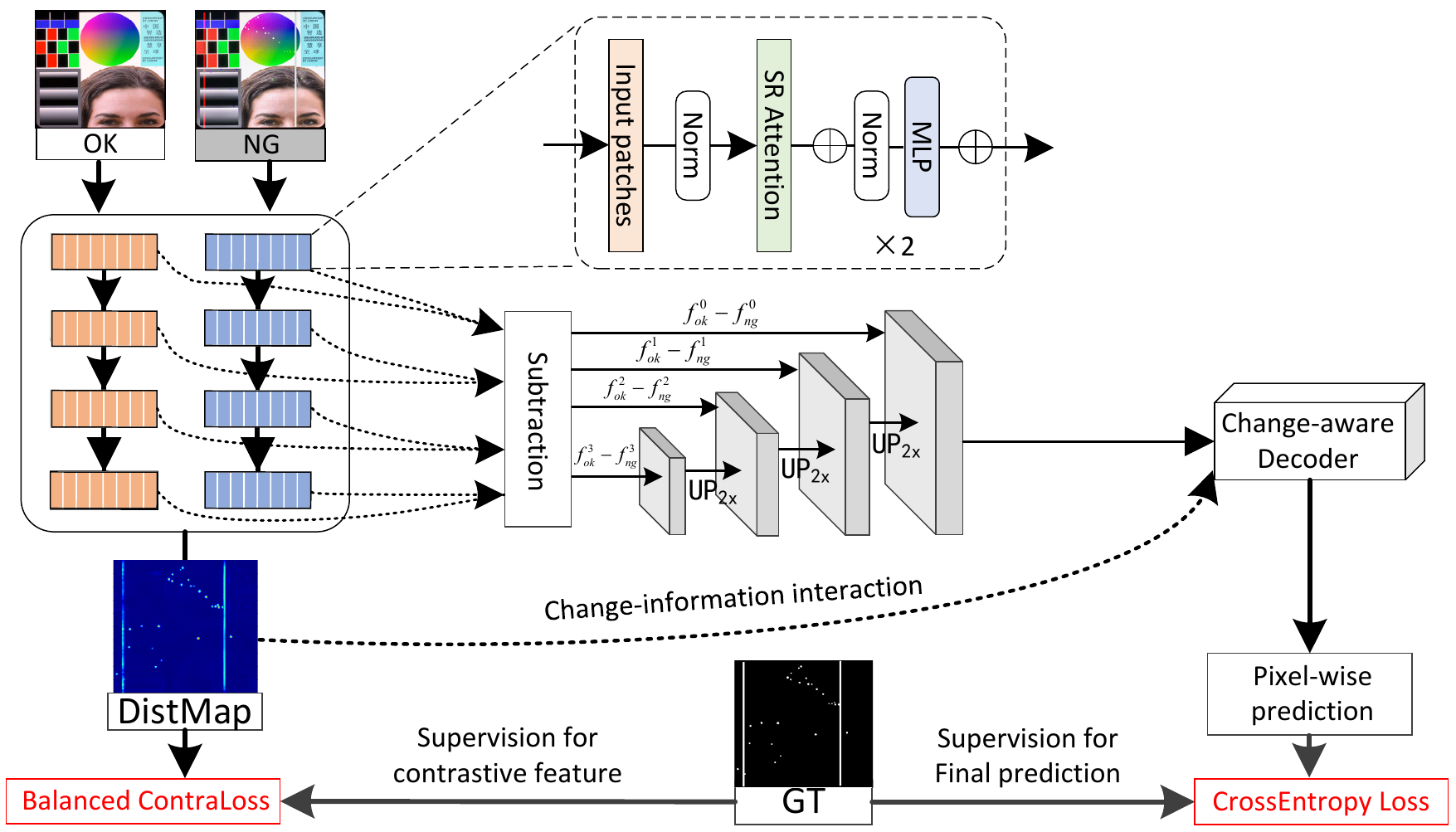}
	\caption{The pipeline of our change-aware Siamese network, which consists of a difference-indicating encoder that extracts contrastive features and a change-aware decoder that applies feature difference (DistMap) to assist in defect localization. The cross-entropy and balanced contrastive loss are adapted for training. }\label{fig3}
\end{figure}

Figure \ref{fig3} depicts our pipeline of change-aware Siamese network. The contrastive encoder extracts deep feature differences between the defective and defect-free samples. The change-aware decoder incorporates change information from the encoder to assist defect localization. The feature distance (DistMap) is used for change information interaction between the encoder and decoder. Specifically, the encoder contains an efficient Transformer-based backbone with four Transformer blocks \cite{vaswani2017attention,SegFormer} using shared weights. Then the hierarchical features are fused via multi-stage subtraction and upsampled to high resolution before decoding. In the decoding stage, the DistMap is used to introduce change information for locating pixel-wise defects. The whole network is supervised by two loss functions, where the cross-entropy loss is used to evaluate the similarity between the predictions and the corresponding ground truth, while the balanced contrastive loss is used to distinguish the features of defective regions from that of defect-free regions.

\subsubsection{Contrastive Feature Encoder}\label{subsubsec3-2-1}

We design an efficient Transformer-based encoder to learn contrastive features with an implicit metric for feature comparison. To improve the efficiency since there are double computation costs for processing paired inputs, we draw the inspiration of sequence reduction attention \cite{wang2021pyramid,SegFormer}, as illustrated in Figure \ref{fig-sub_modules} (a). 
A major bottleneck of the vanilla self-attention mechanism \cite{vaswani2017attention} is the quadratic complexity with long sequence inputs, which is formulated as:
\begin{equation}
	\text{Attention}(Q,K,V) = \text{Softmax}(\frac{QK^T}{\sqrt{d_{k}}})V.
\end{equation}
where matrices $Q, K,$ and $V$ have the same dimensions $N\times C$, and $d_k=N$. We adopt the ratio $R$ to reduce the length of sequence $K$ as follows:
\begin{equation}
	\hat{K} = \text{reshape}(\frac{N}{R},C\cdot R)(K)
\end{equation}
\begin{equation}
K = \rm{linear}(C\cdot R, C)(\hat{K})
\end{equation}
 where the sequence $K$ is initially reshaped to $\frac{N}{R} \times C \cdot R$, followed by a linear layer that processes a sequence of shape $(C \cdot R)$ and produces a $C$-dimensional sequence. Consequently, the dimensions of the new $K$ become $\frac{N}{R} \times C$, effectively reducing the complexity of the self-attention process from $O(N^2)$ to $O\left(\frac{N^2}{R}\right)$.
 Each sequence reduction attention (SRA) module comprises a residually connected sequence reduction attention unit and a multi-layer perceptron (MLP). We employ two SRA modules at each Transformer stage, assigning reduction ratios of [8, 4, 2, 1] for the four stages, respectively.
 
The hierarchical Transformer blocks encode the defective and defect-free images in parallel using shared weights since the image pairs differ only in minimal defective regions.  
Denoting the pyramid features as $\{f_m^n|m={0,1},n={0,1,2,3}\}$, where $m$ indicates the two Siamese branches, and $n$ denotes the four feature layers. The feature distance at position $(i,j)$ is
\begin{equation}
	\begin{split}
	\text{DistMap}\left(i,j\right) &= \left\Vert f_{}^\text{ng}\left(i,j\right)-f_{}^\text{ok}\left(i,j\right)\right \Vert_{2},\\
 	f_{}^\text{ng}&=\rm{concat}{({f_0^1,f_0^2,f_0^3,f_0^4})},\\	f_{}^\text{ok}&=\rm{concat}{({f_1^1,f_1^2,f_1^3,f_1^4})},
	\end{split}
\end{equation}
where $f^{ng}$ and $f^{ok}$ denote the features from defective and defect-free images, respectively. The contrastive loss (CL) is formulated as 
\begin{equation}
	\text{CL}=\left\{ \begin{matrix}
		\text{DistMap}\left( i,j \right)-{\tau }_{ok},\quad\quad &y(i,j)=0,  \\
		\max \left( 0,{\tau }_\text{ng}-\text{DistMap}\left(i,j \right) \right),\quad &y(i,j)=1,  \\
	\end{matrix}\right.
\end{equation}
where $y(i,j)$ is the ground truth, with values 0 or 1 indicating whether the point is unchanged or changed, respectively. ${\tau}_\text{ok}$ and ${\tau}_\text{ng}$ are non-negative thresholds. When $y(i,j)=0$ (i.e., unchanged point), the feature distance is expected to reduce towards ${\tau}_\text{ok}$, which is close to 0. Conversely, when $y(i,j)=1$ (i.e., changed point), the feature distance is encouraged to increase towards ${\tau}_\text{ng}$. We set the ${\tau}_\text{ng}$ and ${\tau}_\text{ok}$ as 2.2 and 0.3 according to \cite{DASNet2021}.

The original contrastive loss is proposed for binary change detection. However, when there is more than one type of defect to be modeled as changed regions (i.e., $y \in {1,2,..,c}$), the sample-amount imbalance between them leads to imbalanced contrastive supervision. Hence, we propose to extend it with a multi-class balanced factor. Given the proportion of certain change categories to the total change areas (i.e., $y(i,j)=1$), the balance factor is defined as
\begin{equation}
	B_{p}=\frac{1}{{f}_{p}}=\frac{1}{{{n}_{p}}}{\sum\limits_{q}^{C}{{{n}_{q}}}}.
\end{equation}
 ${f}_{q}$ is the ratio of class $q$ sample points to the total number of change sample points, where $n_q$ and $n_p$ denote the number of points in class $q$ and class $p$, respectively. The balanced contrastive loss (BCL) can be defined as
\begin{equation}
	\text{BCL}=\left\{ \begin{matrix}
		& \text{CL}, & \ y(i,j)=0, \\ 
		& \sum\limits_{{c}^{l}=0}^{C}{{B}_{l}}, \cdot \text{CL}({y(i,j)}=c^l) & \ y(i,j)\in{1,2,..,C}.  
	\end{matrix}\right.
\end{equation}
It places greater emphasis on less common change categories, resulting in a well-balanced distribution of loss across different types of changes.

\subsubsection{Change-Aware Decoder}\label{subsubsec3-2-2}
\begin{figure}[th]%
	\centering
	\includegraphics[width=0.9\textwidth]{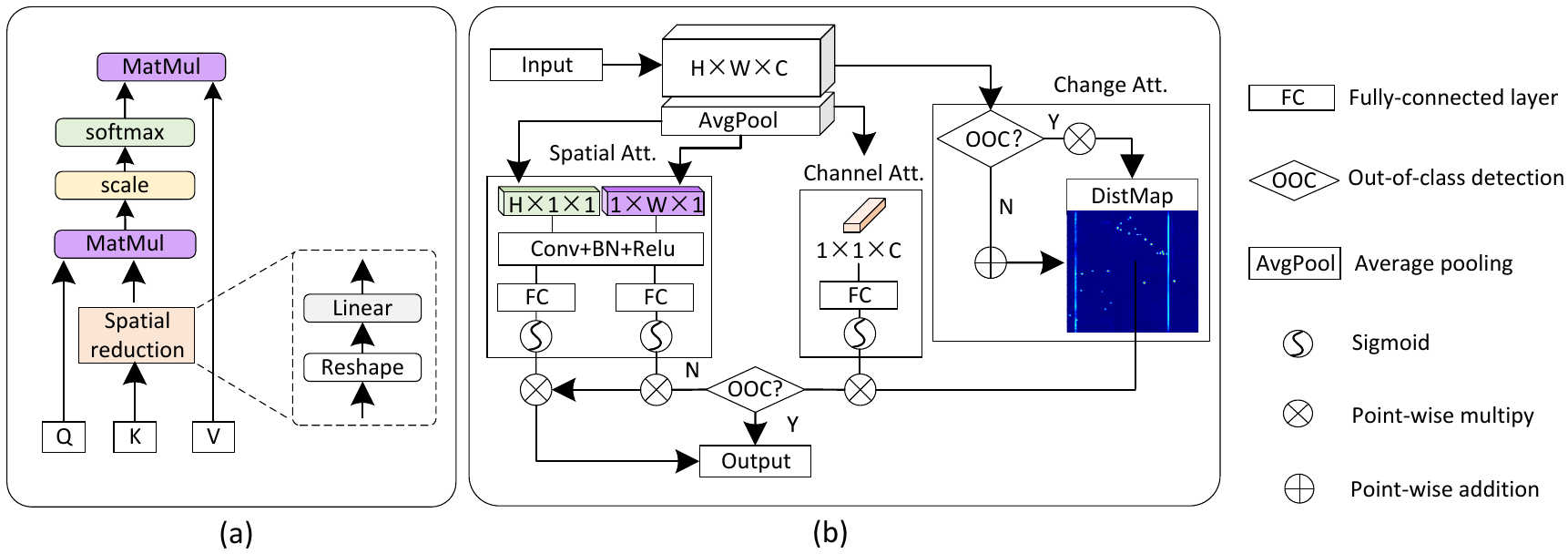}
	\caption{The basic modules. (a) The sequence reduction attention utilizes the spatial reduction layer to reduce the complexity of the self-attention module from $O(N^2)$ to $O\left(\frac{N^2}{R}\right)$. (b) The change-aware decoder, based on a 3-dimensional (horizontal, vertical, and depth) attention module, utilizes the distMap carrying change information in different ways when detecting intra-class and OOC objects.} \label{fig-sub_modules}
\end{figure}

The attention mechanism is widely applied to model contextual information. However, the arbitrary location distribution and weak association with the surroundings of defects have seriously corrupted the spatial context. To this end, we proposed a novel change attention mechanism named change-aware decoder (CAD), which introduces change information to assist in the location of the defect objects. Specifically, the feature difference obtained from the contrastive feature encoder is skip-connected to the decoder and plays different roles when detecting intra-class or OOC objects. The structure of CAD is shown in Figure \ref{fig-sub_modules} (b).

Initially, we extend the lightweight coordAttention \cite{hou2021coordinate} into a 3-dimensional attention module, which allows us to achieve considerable precision in feature decoding while maintaining a low parameter cost.
Constrained by the balanced contrastive loss, the DistMap exhibits high activation values for the change region and low values for the constant region. Current semantic segmentation methods have proven effective when detecting intra-class defects with a known appearance. Hence, the feature difference is added to the encoded features to assist in locating defects, which is
\begin{equation}
	\text{output}=\operatorname{ChangeAtt}(\text{input} + \text{distMap}),
\end{equation}
where $'+'$ means bit-wise summation, and $\operatorname{ChangeAtt}$ here is the combination of channel Attention (CA), horizontal attention (HA), and vertical attention (VA). 
The $\operatorname{ChangeAtt}$ is derived from
\begin{equation}
	\operatorname{ChangeAtt}( \cdot ) = \text{CA}( \cdot ) \otimes \text{HA}( \cdot )\otimes \text{VA}( \cdot ),
\end{equation}
where $\otimes$ means element-wise multiplication. 
However, when encountering OOC defects with unknown appearances (for instance, training with line defects and testing with point defects), the reliance on defect appearance becomes ineffective. In fact, it could be argued that when defect patterns are modeled too accurately on the training set, it may lead to poorer generalization performance on the test set. In such scenarios, change information becomes the primary indicator for defect localization. Consequently, the DistMap interacts with the encoded features multiplicatively after normalization ($\operatorname{Norm}$) to aid in this process, which is
\begin{equation}
	\text{output}=\operatorname{ChangeAtt}(\text{input}),
\end{equation}
\begin{equation}
	\operatorname{ChangeAtt}( \cdot ) =\text{CA}( \cdot ) \otimes \operatorname{Norm}(\text{distMap})\otimes( \cdot ),
\end{equation}
In this context, the multiplication operation incorporates a robust prior to specifically target the change regions. The distMap serves as a spatial context prior, replacing the conventional horizontal or vertical attention mechanisms. Its purpose is to guide the model in identifying potential defects within the change areas. Notably, Figure \ref{fig_ablation} demonstrates that the distMap provides a coarse representation of the final outcome, with the so-called defective regions aligning precisely with the actual regions of change.

\subsubsection{Loss Function}\label{subsubsec3-2-3}

The BCL and cross-entropy loss are employed for training the network. The BCL guides the model to learn contrastive features as mentioned in section \ref{subsubsec3-2-1}. The cross-entropy loss for a single point $(i,j)$ is defined as
\begin{equation}
	\text{CEL}=-\log \frac{{{e}^{\hat{y}(i,j,{{c}^{y}})}}}{\underset{c^{k}=0}{\overset{C-1}{\mathop{\sum }}}\,{{e}^{\hat{y}(i,j,c^{k})}}},
\end{equation}
where ${c}^{y}$ is the true category of a sample point, $C$ is the total categories, and $\hat{y}(i,j,c^k)$ indicates the predicted probability of class $c^k$. 

When detecting intra-class defects, we employ the Cross-Entropy Loss (CEL) and the BCL simultaneously. In situations where the defect appearance remains uncertain, the change information captured by BCL becomes the primary basis for defect localization. The overall loss function used during model training is as follows:
\begin{equation}
	\rm{loss}=\begin{cases}
		\lambda _{1} \text{CE}+\lambda_{2} \text{BCL} & \ C_\text{train} =C_\text{test},\\
		\text{BCL} & \ C_\text{train} \neq C_\text{test}.
	\end{cases}
\end{equation}
where $C_\text{train}$ and $C_\text{test}$ are the set of defect categories in the training and testing phases, respectively. $\lambda_1$ and $\lambda_2$ are set to 1 in our experiment. 

\section{Experiments and Results}\label{sec4}
\subsection{Datasets}\label{subsec3-1}
Three datasets are involved for evaluation, including our synthetic LCD and the PKU-Market-PCB \cite{ding2019tdd} datasets, which are charaterized by the complex background and tiny texture anomalies. Additionally, the anomaly detection benchmark MVtec-AD \cite{bergmann2019mvtec} is used for validating the generizibility of our method. 

\begin{figure}[!th]%
	\centering
	\includegraphics[width=0.8\textwidth]{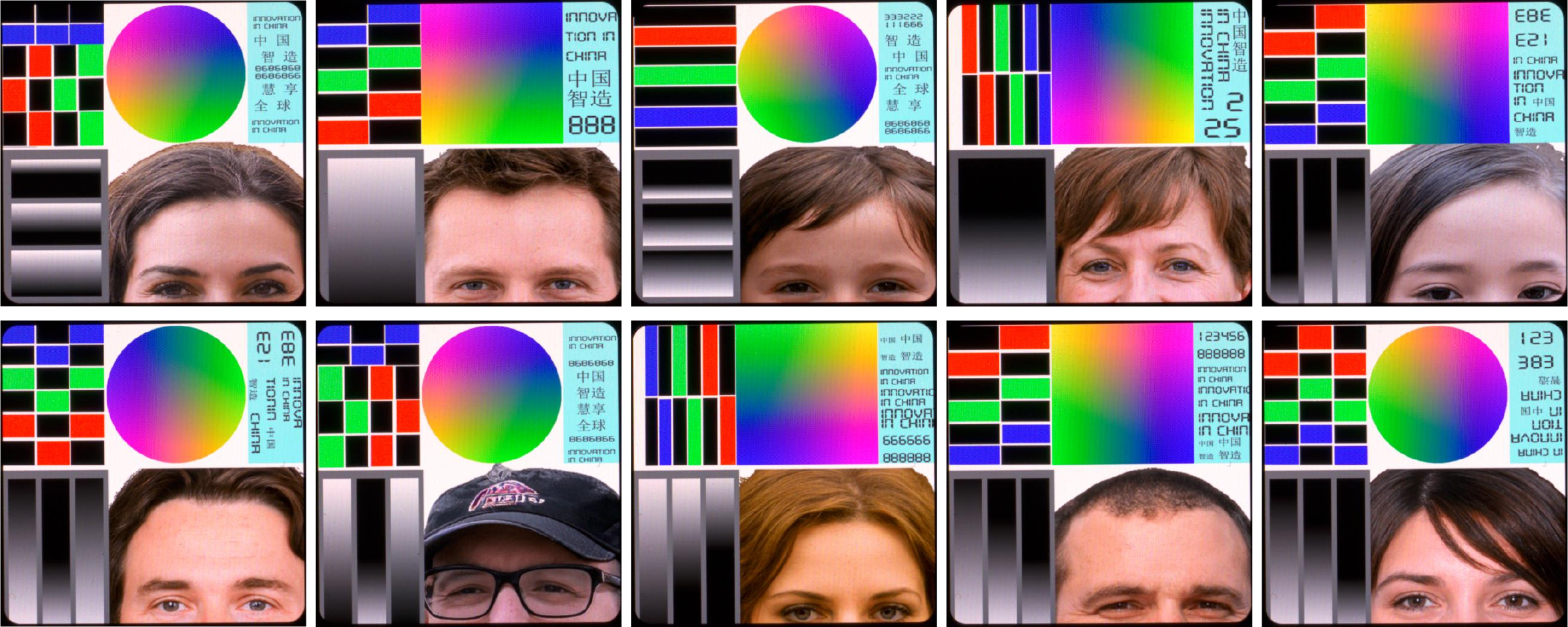}
	\caption{Ten defect-free LCD display patterns. In real inspection process, the industrial LCD display patterns are constructed with RGB blocks, gray transition, color maps, characters, and faces to reveal various types of defects (e.g, point, line, and Mura defects \cite{ming2021survey}).}\label{fig1}
\end{figure}

{\bf Synthetic LCD defect dataset}. 
To validate our model's capability in segmenting defect in various of imaging, production conditions, and defect appearances, we constructed a synthetic LCD defect dataset termed SynLCD. During the real-world LCD inspection process, some specific display patterns are designed to reveal various types of defects (e.g, point, line, and Mura defects \cite{ming2021survey}). These patterns are constructed with pure color blocks, color maps, text blocks, grayscale transitions, and human faces. Figure \ref{fig1} has depicted 10 defect-free display patterns. 

\begin{figure}[!h]%
	\centering
	\includegraphics[width=0.8\textwidth]{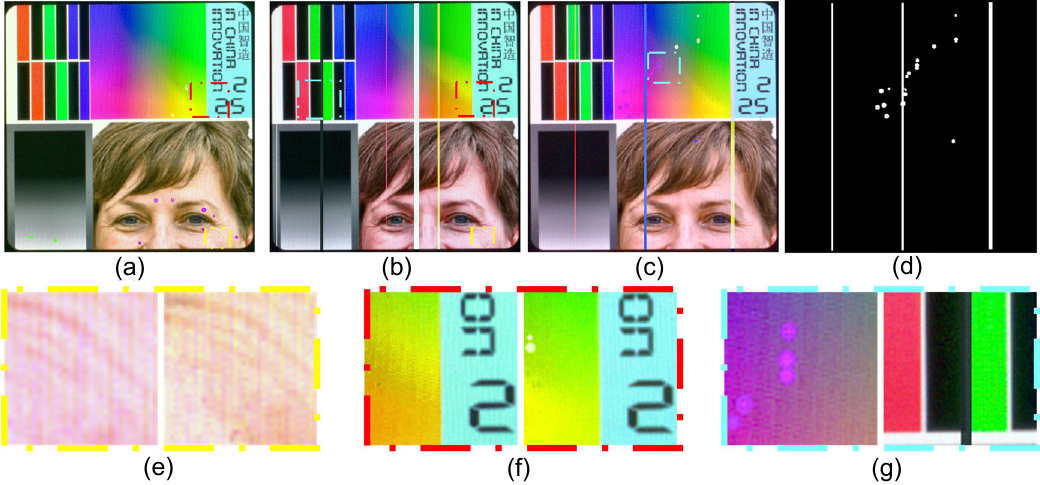}
	\caption{Samples of SynLCD and the dataset challenges. (a) Abnormal points defect sample; (b) line defect sample; (c) mixed defect sample; (d) binary label of mixed defect image. (e) RGB deviation and irregular screen texture; (f) nonlinear saturation difference. (g) low contrast abpt and line defects.}\label{fig2}
\end{figure}

The synLCD dataset includes 3 types of defect samples with random positions and distribution: line defects, abnormal points (abpt), and mixed defects, as presented in Figure \ref{fig2}.  Some of these defects closely resemble the background patterns. For line defects, they exhibit low contrast with the background, spanning across the entire screen.

\begin{table}[tbp]
    \renewcommand\arraystretch{0.8}
	\caption{Statistical details about SynLCD dataset.}
	\label{table1}
	\vspace{2mm}
	\centering
	\scalebox{0.75}{
	\begin{tabular}{cccc}
		\toprule
		Attributes & Type & Values & Remarks \\
		\midrule
		\multirow{4}*{Amount} & background pattern & 10 & variation in face and color-map, etc.\\
		~ & Defect types & 3 & line, abpt and mixed defects \\
		~ & Defect Samples & 10$\times$300$\times$3 & 300 samples for each type and each pattern \\
		~ & Nondefect Samples & 10$\times$900 & variation in brightness and contrast, etc. \\
		\cmidrule(r){2-4}
		\multirow{3}*{Defect} & shape & 2 & line and abpt\\
		~ & color & 5 & black, white, red, green, blue \\
		~ & opacity & 10$\%$-100$\%$ & 10$\%$ interval \\
		~ & width & 3-33 pixels & 3$\%$ interval \\
		\cmidrule(r){2-4}
		\multirow{3}*{Screen} & brightness & bias: 1-6 & 1 interval\\
		~ & contrast & alpha: 0.5-1.5 & 0.1 interval\\
		~ & ISO noise & 10$\%$-100$\%$ & 10$\%$ interval \\
		~ & RGB deviation & 3-33 grayscale & 3 interval \\
		\bottomrule
	\end{tabular}}
\end{table}

Table \ref{table1} shows the statistical details of SynLCD. According to the assumption in Eq. \eqref{eq1}, the defect image $x_\text{ng}$ is formed by superimposing a clean surface image $x_\text{ok}$ with the defect $x_\text{defect}$ after applying a non-linear overall surface change $\sigma$. To generate line defects, we first divide the clean image into (K) areas. Next, in each area, we pre-draw a line with random color, transparency, and width. These lines traverse the screen, simulating real-world line defects.
Abnormal points tend to appear in high-frequency transition regions such as edges, hair, and text. To create abpt samples, we vary the grayscale threshold from 50 to 200 to obtain segmentation results at each threshold. From these segmentation results, we extract a set of edge points. Subsequently, we randomly cluster these points using K-means clustering, assigning each subclass a random color, scale, and transparency.
Once we obtain both types of defects, we overlay them onto the clean image using Gaussian blur and Poisson seamless fusion \cite{2003Possion}. This process introduces random luminance, contrast, ISO noise, and RGB color bias, enhancing sample diversity. To prevent sample imbalance interference during the classification task, we generate 300 defective and defect-free samples for each clean image in Figure \ref{fig1}.
In total, there are 4,200 training samples (7 standard background patterns and 600 samples for each pattern) and 1,800 testing samples.  

{\bf PKU-Market-PCB}. The PKU-Market-PCB dataset\footnote{https://robotics.pkusz.edu.cn/resources/dataset/} comprises 1,386 images along with 6 types of defects to validate the generalizability of our model in the scene of complex background and tiny defects. The original images exhibit inconsistent sizes. To streamline the training process, we resized and cropped the original images into $1000\times1000$ sub-images, retaining only those containing defects. Finally, there are 1,566 (70\%) images for training and 676 (30\%) images for testing. The preprocessed PCB dataset is included with our source code for accessibility\footnote{https://github.com/qaz670756/CADNet}. 

{\bf MvTec-AD.} To further validate our model in detecting general defects, we conduct comparison in the MvTec-AD \cite{bergmann2019mvtec}. It is a widely used anomaly detection benchmark. To facilitate more effective training and achieve precise defect segmentation, we reorganized the original dataset for fully-supervised training. The original 5,354 images, along with their corresponding ground-truth annotations, were randomly shuffled and divided into two subsets: 3,747 (70\%) images for training and 1,607 (30\%) images for testing.

\subsection{Experiment setting and metrics}
{\bf Implementation details}. Our model is realized with mmsegmentation \footnote{https://github.com/open-mmlab/mmsegmentation} and trained with an RTX3090 GPU. The input images of SynLCD are resized into $512\times512$ with common data augmentations including random crop, flip, and color normalizing during training. All models are trained for 30 epochs (i.e. 126,000 iterations). In the context of semi-supervised learning, we vary the proportion of labeled samples between 0\%, 5\%, 10\%, and 15\%. Due to the diverse numbers of training samples, we maintain a fixed iteration count of 126,000 when exclusively using labeled samples in the target set. To compare with UAPS \cite{sime2023uncertainty}, which utilizes unlabeled data for training, we follow the established setting in \cite{sime2023uncertainty} by incorporating 10\% of unlabeled data.

{\bf Metrics}. We involve the semantic segmentation metrics for evaluating the pixel-wise defect predictions, including mean Intersection over Union (mIoU), Accuracy (Acc), and Fscore as also denoted in \cite{deeplabv3plus2018,liu2021zoominnet}. Defining TP, FP, and FN as abbreviations for True Positive, False Positive, and False Negative, respectively.
The metrics are outlined as follows:
\begin{itemize}
	\setlength{\itemsep}{1pt}
	\setlength{\parsep}{1pt}
	\setlength{\parskip}{1pt}
	\item precision (P) and recall (R): ${\small \text{TP/(TP+FP)}, \text{TP/(TP+FN)}}$,
	\item Fscore: {\small $\text{2PR/(P+R)}$},
	\item accuracy (Acc): {\small ${\text{TP}+\text{TN}}/(\text{TP}+\text{FN}+\text{FP}+\text{FN})$},
	\item mIoU: $\frac{1}{C} \sum_{i=0}^{(C-1)} \frac{\text{TP}_i}{\text{TP}_i+\text{FP}_i+\text{FN}_i}$.
\end{itemize}
To measure model complexity, we use parameters (Params) and Giga floating point of operations (GFLOPs). Tn all tables, the up-arrow means the higher the better, while the down-arrow means the lower the better.

{\bf Compared methods}. Our model is evaluated from two aspects: (1) The intra-class segmentation performance aims to demonstrate the superiority of change modeling over appearance modeling when there are pixel-wise labels available. Six semantic segmentation methods are involved for comparison as shown in Table \ref{table_compared_methods}. (2) The out-of-class segmentation aims to evaluate the model robustness facing class shift as defects in a real-world production environment would not have a consistent appearance. Five SOTA semi-supervised methods are involved for comparison as given in Table \ref{table_compared_methods}.

\begin{table}[!b]
\renewcommand\arraystretch{1.0}
	\caption{An overview of fully-supervised and semi-supervised segmentation methods for comparison.}
	\label{table_compared_methods}
	\vspace{2mm}
	\centering
    \scalebox{0.63}{
	\begin{tabularx}{1.6\textwidth}{X|X}
		\toprule
		\textbf{Fully-Supervised Methods} & \textbf{Semi-Supervised Methods}\\
		\hline
			{\bf FCN} \cite{2015FCN}: utilizes fully convolutional layers to realize dense prediction for arbitrary-sized images. &  {\bf DCT} \cite{qiao2018deep}: employs one network to ensure consistency across different p views of a given sample.  \\
            {\bf PSPNet} \cite{pspnet2017}: Utilizes global context aggregation through pyramid pooling for complicated scene parsing. & {\bf CPS} \cite{chen2021semi}: enforces consistency between two segmentation networks initialized differently.    \\
            {\bf DeepLabV3+} \cite{deeplabv3plus2018}: introduced the atrous spatial convolutional pyramid (ASPP) to enhance the multi-scale contextual information. & {\bf UAMT} \cite{yu2019uncertainty}: encourages consistent predictions under different perturbations and estimates uncertainty to learn from unlabeled data.   \\
            {\bf DANet} \cite{fu2019dual}: enhances segmentation by adaptively integrating semantic dependencies in spatial and channel dimensions via the self-attention mechanism.& {\bf UCC} \cite{fan2022ucc}: employs a shared encoder with dual decoders and enforces consistency between the decoders with data augmentations. \\
            {\bf OCRNet} \cite{OCRNet2020}: introduces object-contextual representations for semantic segmentation, leveraging pixel-object relationships to augment pixel representations. & {\bf UAPS} \cite{sime2023uncertainty}: dynamically blends pseudo-labels from multi-head outputs during a single forward pass for uncertainty regularization.\\
            {\bf SegFormer} \cite{xie2021segformer}: presents a streamlined semantic segmentation framework by integrating Transformers with lightweight MLP decoders.   \\
		\bottomrule
	\end{tabularx}}
\end{table}

\subsection{Quantitative Results and Comparison}
\subsubsection{Fully-supervised segmentation}

In this section, we compare our proposed method with the fully-supervised models in the aspects of intra-class and out-of-class segmentation performance.
From the results of Table \ref{table_fully_seg_compare}, our model achieves a remarkable improvement over the other segmentation models. Specifically, our model exhibits improved performance across the four metrics ($\rm{IoU_{line}}$, $\rm{IoU_{abpt}}$, $mIoU$, $mFscore$) by 12.65\%, 0.82\%, 8.17\%, and 4.15\%, compared to the runner-up results. In Table \ref{table_fully_seg_compare_pcb} and \ref{table_fully_seg_compare_mvtec}, our model obtains the best outcomes across all metrics in the PCB and MvTec-AD datasets.

\begin{table}[!t]
	\renewcommand\arraystretch{1.1}
	\caption{Comparison with the SOTA semantic segmentation methods in SynLCD dataset. {\color{red}{Red}}, {\color{green}{green}} and {\color{blue}{blue}} indicate the top three results for each metric.}
	\label{table_fully_seg_compare}
	\vspace{2mm}
	\centering
	\scalebox{0.65}{
		\begin{tabular}{l|ccccc|cc}
			\toprule
			Method& $IOU_\text{line}\uparrow$  & IOU$_\text{abpt}$$\uparrow$& mIOU$\uparrow$&   mAcc$\uparrow$&   mFscore$\uparrow$  & MParams$\downarrow$ & GFLOPs$\downarrow$\\
			\midrule
			FCN \cite{2015FCN}&51.86& 11.48& 31.67& 36.06& 44.45& 49.5& 57.91 \\
			PSPNet  \cite{pspnet2017}& 79.00& 52.54& 65.77& 71.56& 78.58&	12.76 &	54.27  \\
			DeepLabV3+ \cite{deeplabv3plus2018}&   {81.96}& \color{green}{72.93}& \color{green}{77.45}& \color{red}{90.24}& \color{green}{87.22}& 43.58&	176.22 \\
			DANet  \cite{fu2019dual} & 79.92& 57.04& 68.48& 76.27& 80.74& 49.82 &199.05 \\
			OCRNet  \cite{OCRNet2020}&\color{green}{83.46}& 62.19& 72.83& \color{blue}{86.08}& 83.84& {12.07}& {52.83} \\
			SegFormer  \cite{xie2021segformer}& \color{blue}{82.99}& \color{blue}{69.62}& \color{blue}{76.31}& {83.68}& \color{blue}{86.39}& {3.72}& {6.37} \\
   \rowcolor{gray!10}
			Our-CADNet & {\color{red}94.02}&	{\color{red}73.53}&	{\color{red}83.78}&	{\color{green}89.05}&	{\color{red}90.84}&	{3.90}&	{8.21}\\
			
			\bottomrule
\end{tabular}}
\end{table}

\begin{table}[!bp]
	\renewcommand\arraystretch{1.1}
	\caption{Comparison with the SOTA semantic segmentation methods in the PCB Dataset. {\color{red}{Red}}, {\color{green}{green}} and {\color{blue}{blue}} indicate the top three results for each metric.}
	\label{table_fully_seg_compare_pcb}
	\vspace{2mm}
	\centering
	\scalebox{0.65}{
		\begin{tabular}{l|ccccccccc}
			\toprule
			Method&   IOU$_{c1}$$\uparrow$& IOU$_{c2}$$\uparrow$&IOU$_{c3}$$\uparrow$&IOU$_{c4}$$\uparrow$&IOU$_{c5}$$\uparrow$&IOU$_{c6}$$\uparrow$& mIOU$\uparrow$&   mAcc$\uparrow$&   mFscore$\uparrow$  \\
			\midrule
			FCN \cite{2015FCN}& 50.13 &	69.19&	68.65&	45.45&	50.35&	36.36&	53.35&	60.80&	68.79
			\\
			PSPNet  \cite{pspnet2017}&74.04	&72.59&	72.61&	71.29&	66.39&	72.46&	71.56&	81.77&	83.40
			   \\
			DeepLabV3+ \cite{deeplabv3plus2018}&75.39&	{\color{green}73.56}&	{\color{green}74.22}&	{\color{blue}73.57}&	69.94&	76.47&	{\color{blue}73.85}&	82.10&	{\color{blue}84.94}
			    \\
			DANet  \cite{fu2019dual} &74.31&	{\color{blue}73.02}&	71.21&	72.14&	68.86&	75.02&	72.42&	82.01&	83.99
			  \\
			OCRNet  \cite{OCRNet2020}&{\color{green}76.08}&	73.00&	{\color{blue}73.78}&	{\color{green}75.98}&	{\color{green}71.13}&	{\color{green}78.13}&	{\color{green}74.68}&	{\color{green}83.45}&{\color{green}85.48}
			 \\
			SegFormer  \cite{xie2021segformer}&{\color{blue}75.79}&	71.39&	72.31&	72.29&	{\color{blue}70.75}&	{\color{blue}78.04}&	73.42&{\color{blue}82.29}&	84.65
			 \\
			\rowcolor{gray!10}
	
			Our-CADNet& {\color{red} 77.21}&	{\color{red} 73.98}&	{\color{red} 75.08}&	{\color{red} 79.95}&	{\color{red} 76.47}&	{\color{red} 82.44}&	{\color{red} 77.52}&	{\color{red} 85.87}&	{\color{red} 87.31}
				
			 \\
			\bottomrule
	\end{tabular}}
\end{table}

In terms of efficiency, our model has comparable parameters to SegFormer, and both surpass other models significantly in computation. Our model shows substantial improvements over SegFormer, with a 1.84 GFLOPs increase resulting in 9.79\% higher mIOU, 6.42\% higher mAcc, and 5.15\% higher mFscore. 
This underscores our model's efficiency, rendering it suitable for deployment in industrial devices with limited computational resources.

\begin{table}[!tb]
	\renewcommand\arraystretch{1.1}
	\caption{Comparison with the SOTA semantic segmentation methods in the MvTec-AD Dataset. {\color{red}{Red}}, {\color{green}{green}} and {\color{blue}{blue}} indicate the top three results. Note that there are 15 classes in MvTec-AD and 6 of them are reported here.}
	\label{table_fully_seg_compare_mvtec}
	\vspace{2mm}
	\centering
	\scalebox{0.65}{
		\begin{tabular}{l|ccccccccc}
			\toprule
			Method&   IOU$_{c1}$$\uparrow$& IOU$_{c2}$$\uparrow$&IOU$_{c3}$$\uparrow$&IOU$_{c4}$$\uparrow$&IOU$_{c5}$$\uparrow$&IOU$_{c6}$$\uparrow$& mIOU$\uparrow$&   mAcc$\uparrow$&   mFscore$\uparrow$  \\
			\midrule
			FCN \cite{2015FCN}& 76.10&	60.14&	35.93&	69.73&	13.51&79.65&58.14&	64.84&	70.00
			\\

      PSPNet \cite{pspnet2017}& 72.00&	{\color{green}68.24}&	43.86&	{\color{red}74.89}&	{\color{blue}42.43}&	{\color{blue}83.44}& {\color{blue}65.42}&	{\color{blue}76.25}&	{\color{green}77.58}
\\

			DeepLabV3+ \cite{deeplabv3plus2018}&{\color{blue}76.65}&	63.48&	41.18&	72.31&	34.93&	81.12&63.77&	{\color{green}77.59}&	76.19

			    \\
			DANet  \cite{fu2019dual} &75.13&	56.37&	37.95&	{\color{blue}72.42}&	27.10&	80.92&61.63&	72.49&	73.94

			  \\
			OCRNet  \cite{OCRNet2020}&70.89	&{\color{blue}65.18}&	{\color{blue}45.67}&	65.47&	35.41&	81.51&59.89&	68.98&	72.31
			 \\
			SegFormer  \cite{xie2021segformer}&{\color{green}81.63}&	64.63	&{\color{green}53.81}&	70.81&	{\color{green}44.14}&	{\color{green}84.71} & {\color{green}65.97}&	71.21&	{\color{blue}77.51}

			 \\
			\rowcolor{gray!10}
	
			Our-CADNet& {\color{red}82.60}&	{\color{red}74.16}&	{\color{red}61.19}&	{\color{green}73.06}&	{\color{red}52.69}&	{\color{red}86.41}&  {\color{red}71.35}&	{\color{red}80.85}&	{\color{red}82.24}

			 \\
			\bottomrule
	\end{tabular}}
\end{table}

\subsubsection{Semi-supervised segmentation}

When defect appearances are clearly defined with ample labeled data, general segmentation models like DeepLabv3+ and SegFormer demonstrate satisfactory performance. However, despite the SynLCD dataset simulating real defects and generating both line and abpt defects, they consistently deviate from real defects. A notable concern is that appearance-based modeling cannot ensure robust generalization in real-world applications. Therefore, we delve deeper into defect segmentation under scenarios of limited or even absent labels (out-of-class segmentation).

\begin{figure}[!t]%
	\centering
	\includegraphics[width=1.0\textwidth]{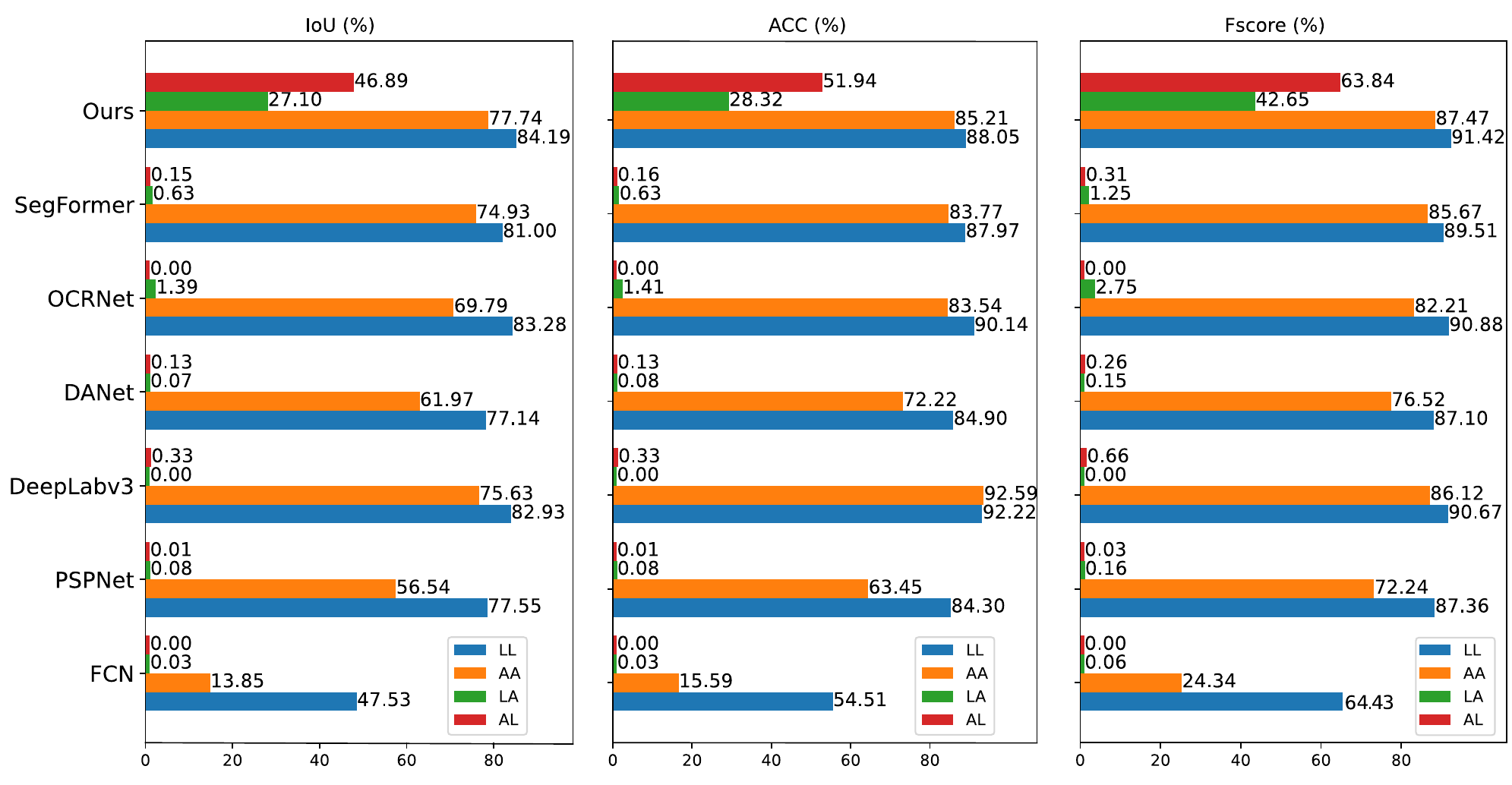}
	\caption{Comparison of cross-testing performance. In this setting, the samples during inference do not appear in the training phase. For LL, AA, LA, and AL, the first character means training with line (L) or abpt (A) set, while the second represents the testing set.}\label{fig6}
\end{figure}

In the series of experiments, denoted as LL, AA, LA, and AL, the first character indicates training on either line (L) or abpt (A), while the second character denotes testing on line (L) or abpt (A).
As results shown in Figure \ref{fig6}, most segmentation models obtain acceptable intra-class segmentation results but fail to detect out-of-class defects (metrics such as IoU, Acc, and Fscore are lower than 0.5\%) due to their appearance-based modeling nature. In contrast, our change-aware model exhibits considerable results when defect appearance is unseen in the training phase. More specifically, measured by the metrics of IoU, Acc, and Fscore, AA (i.e. trained and tested on abpt defect) are 69.77\%, 84.19\%, and 82.20\% respectively, while AL (i.e. trained on abpt and tested on line defect) maintains considerable performance level with 40.5\%, 52.01\%, and 57.66\%, respectively. 
Regarding LA (i.e. trained online and tested on abpt defect), there's a notable decrease in accuracy. It is conceivable that the abpt defects are harder to distinguish from the background with smaller sizes.
 Using synthetic data on the production line can be instrumental in initializing a streamlined model for rough inspection processes, significantly reducing data collection and labeling costs. 
 
 \begin{table}[!t]
 	\renewcommand\arraystretch{0.95}
 	\caption{Comparison with the SOTA semi-supervised segmentation methods in the SynLCD dataset across varying proportions of labeled data (from 0\% to 15\%). 
All models are pre-trained on the abpt defects and subsequently fine-tuned and tested using the line defects. The {\bf bold} font indicates the best results.}
	\vspace{2mm}
\centering
\scalebox{0.8}{
 	\begin{tabular}{l|cccccccc}
 		\toprule
 		\multirow{2}*{Method} & \multicolumn{4}{c}{mIoU$\uparrow$}  & \multicolumn{4}{c}{Fscore$\uparrow$} \\ [-1mm]
 		\cmidrule(r){2-5}
 		\cmidrule(r){6-9}
 		~ & 0\% & 5\% & 10\% & 15\% & 0\% & 5\% & 10\% & 15\% \\
\hline
 		DCT \cite{qiao2018deep} & 0.05 & 56.96 & 73.67 & 71.85 & 0.10 & 71.27 & 84.57 & 82.75 \\
	
 		UAMT \cite{yu2019uncertainty} & 0.44 & 61.68 & 68.73 & 71.96 & 0.88 & 75.48 & 80.94 & 83.15 \\
 	
 		CPS \cite{chen2021semi}& 1.09 & 65.07 & 65.63 & 76.02 & 2.15 & 78.29 & 78.70 & 85.68 \\
 	
 		UCC \cite{fan2022ucc}& 0.015 & 61.40 & 70.48 & 71.55 & 0.03 & 75.41 & 82.27 & 82.78 \\
 	
 		UAPS \cite{sime2023uncertainty} & 0.44 & 58.86 & 74.43 & 81.34 & 0.88 & 72.52 & 84.35 & 89.22 \\
 		\rowcolor{gray!10}
 		
 		Our-CADNet & {\bf 46.89}&	{\bf 82.93}&	{\bf 84.52}&{\bf 84.71}&	{\bf 63.84}&	{\bf 90.87}&	{\bf 91.64}&	{\bf 91.72}\\
 	
 	\bottomrule
 	\end{tabular}}
 	
 	\label{tab:my_label}
 \end{table}

Table \ref{tab:my_label} demonstrates our model's superior performance to five SOTA semi-supervised segmentation methods across different supervision settings. Particularly notable is the fact that when all models are pre-trained solely with abpt defects, only our model achieves satisfactory results, while the others yield collapsed outcomes in the line defects.

\subsection{Ablation studies.}

\begin{figure}[!h]%
	\centering
	\includegraphics[width=1.0\textwidth]{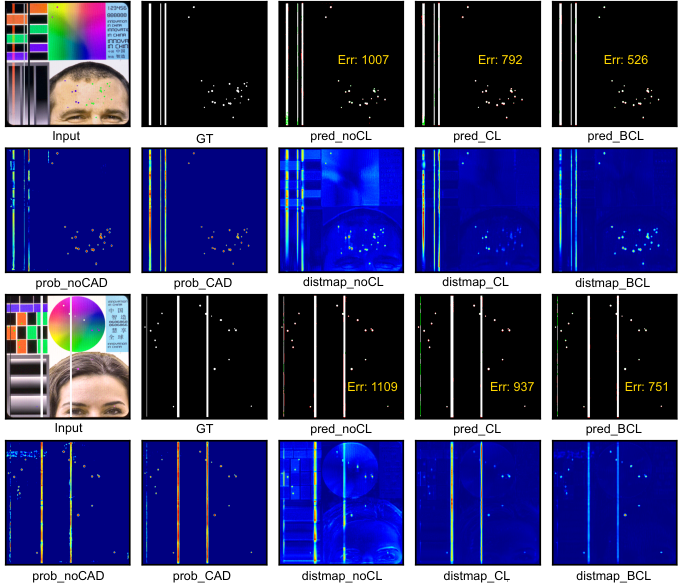}
	\caption{Visual ablation results. It shows the final predictions (pred), probability map (prob) before output and DistMap with or without CAD, CL, and BCL.}\label{fig_ablation}
\end{figure}

In this section, we investigate how the contrastive loss (CL), balanced contrastive loss (BCL), and change-aware decoder (CAD) influence the model. According to the results in Table \ref{table4} and Figure \ref{fig_ablation}, the following conclusions can be drawn:

\begin{table}[!b]
	\renewcommand\arraystretch{1.1}
	\caption{Ablation study about the loss function and decoder. From left to right are cross-entropy loss (CEL), contrastive loss (CL), balanced contrastive loss (BCL), and change-aware decoder.}
	\label{table4}
	\vspace{2mm}
	\centering
	\scalebox{0.75}{
	\begin{tabular}{cccc|ccccccc}
		\toprule
		\bf{CEL}& \bf{CL}& \bf{BCL}& \bf{CAD}& IoU$_{line}$$\uparrow$& IoU$_{abpt}$$\uparrow$ &mIoU$\uparrow$&   mAcc$\uparrow$ &mFscore$\uparrow$& Params$\downarrow$& GFLOPs$\downarrow$ \\
		\midrule
		\checkmark&&&&84.21&	73.00&	78.61&	85.09&	87.91&	3.72&	8.16
		\\
		
		\checkmark&\checkmark &&&89.40&	70.17&	79.78&	85.22&	88.43&	3.72&	8.16
		 \\ 
		\checkmark&&\checkmark  &&89.56&	72.96&	81.26&	87.32&	89.43&	3.72&	8.16
		 \\
            
		\checkmark&&\checkmark &\checkmark &94.02&	73.53&	83.78&	89.05&	90.84&	3.90&	8.21
		 \\
		\bottomrule
	\end{tabular}}
\end{table}

\begin{itemize}
	\setlength{\itemsep}{0pt}
	\setlength{\parsep}{0pt}
	\setlength{\parskip}{0pt}
	\item Leveraging CL to supervise intermediate layers has led to notable improvements in most accuracy metrics without introducing extra computational costs. Visual comparison between distMap\_noCL and distMap\_CL in Figure \ref{fig_ablation} highlights how the contrastive constraint aids in reducing background noise and identifying more discriminative change (defective) regions. Furthermore, distMap\_noCL illustrates that lines are more discernible than abpt regions, indicating an imbalanced contrastive constraint.	
	\item As depicted in distMap\_CL and distMap\_BCL in Figure \ref{fig_ablation}, BCL effectively amplifies the intensity of abpt defects, leading to a further improvement in $\text{IoU}_\text{abpt}$ while maintaining stable $\text{IoU}_\text{line}$. Consequently, there is an overall increase in mIoU and mFscore.	
	\item The CAD model yields enhancements across all accuracy metrics with a minimal increase of less than $0.18M$ parameters and a burden of only $0.05$ GFLOPs. The analysis of prob\_noCAM and prob\_CAM reveals the significance of change information and spatial context in effectively restoring broken lines while mitigating noise detections.
\end{itemize}

\subsection{Qualitative results}

\begin{figure}[!b]%
	\centering
	\includegraphics[width=1.0\textwidth]{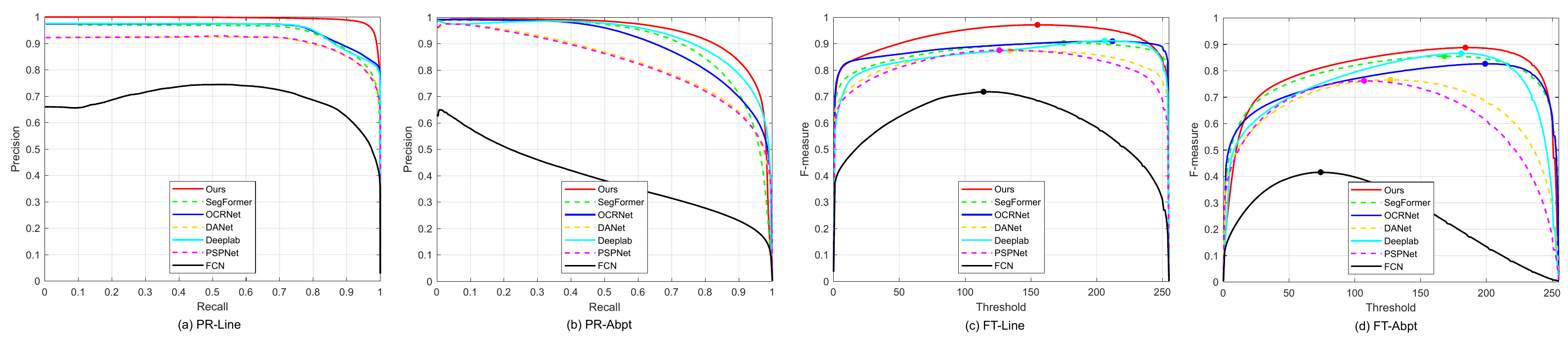}
	\caption{Comparison though precison-recall (PR) and Fscore-threshold (FT) curves. From left to right, the PR curves of the line, the PR curves of the abpt, the FT curves of the line, and the FT curve of the abpt defects.}\label{fig_curves}
\end{figure}

In the left two panels of Figure \ref{fig_curves}, the Precision-Recall (P-R) curves demonstrate that our change-aware network consistently outperforms others, particularly at higher recall values, for both the line and abpt defects. Examining the Fscore-Threshold (FT) curves in the right two panels, our model consistently achieves a higher Fscore across various binary threshold values. Furthermore, the detection of larger-sized line defects generally results in higher precision and Fscore compared to abpt defects.

\begin{figure}[!t]%
	\centering
	\includegraphics[width=1.0\textwidth]{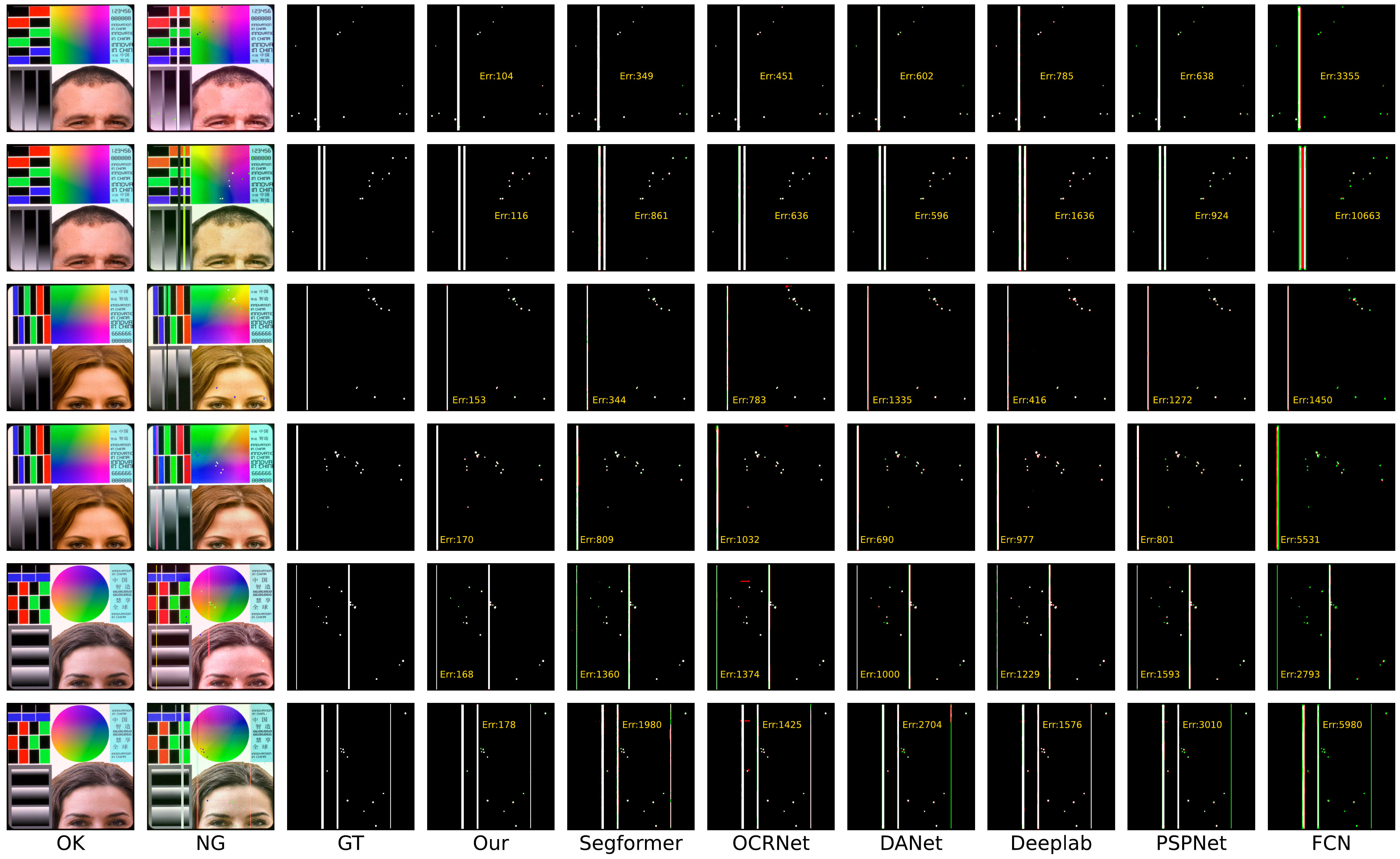}
	\caption{Visual comparison in SynLCD dataset. White color represents the line and abpt defects, while {\color{green}green} color represents missed detections and {\color{red}red} color wrong detections. The errors (Err) in yellow summarise the missed and wrong detections.}\label{fig5}
\end{figure}

Figure \ref{fig5} and \ref{fig_vis_pcb} present further visualization comparison in the SynLCD and PCB datasets. For an intuitive observation, the line and abpt defects are all set to white color: green color denotes missed detections and red color denotes wrong detections. The errors in yellow summarise the missed and wrong detections.
In general, lines are harder to detect completely than abpts because they span over the entire image, which requires the network to model global context over long distances. Thin lines, in comparison, are more likely to be missed than thick lines, as the downsampling during feature extraction may cause information loss. Overall, the FCN is the least effective, as reflected by its accuracy metrics. It has a large number of misses and wrong detections on all the tested images. In contrast, our model outperforms the other methods on all test images significantly fewer parameters and lower computational cost. 

\begin{figure}[!t]%
	\centering
	\includegraphics[width=1.0\textwidth]{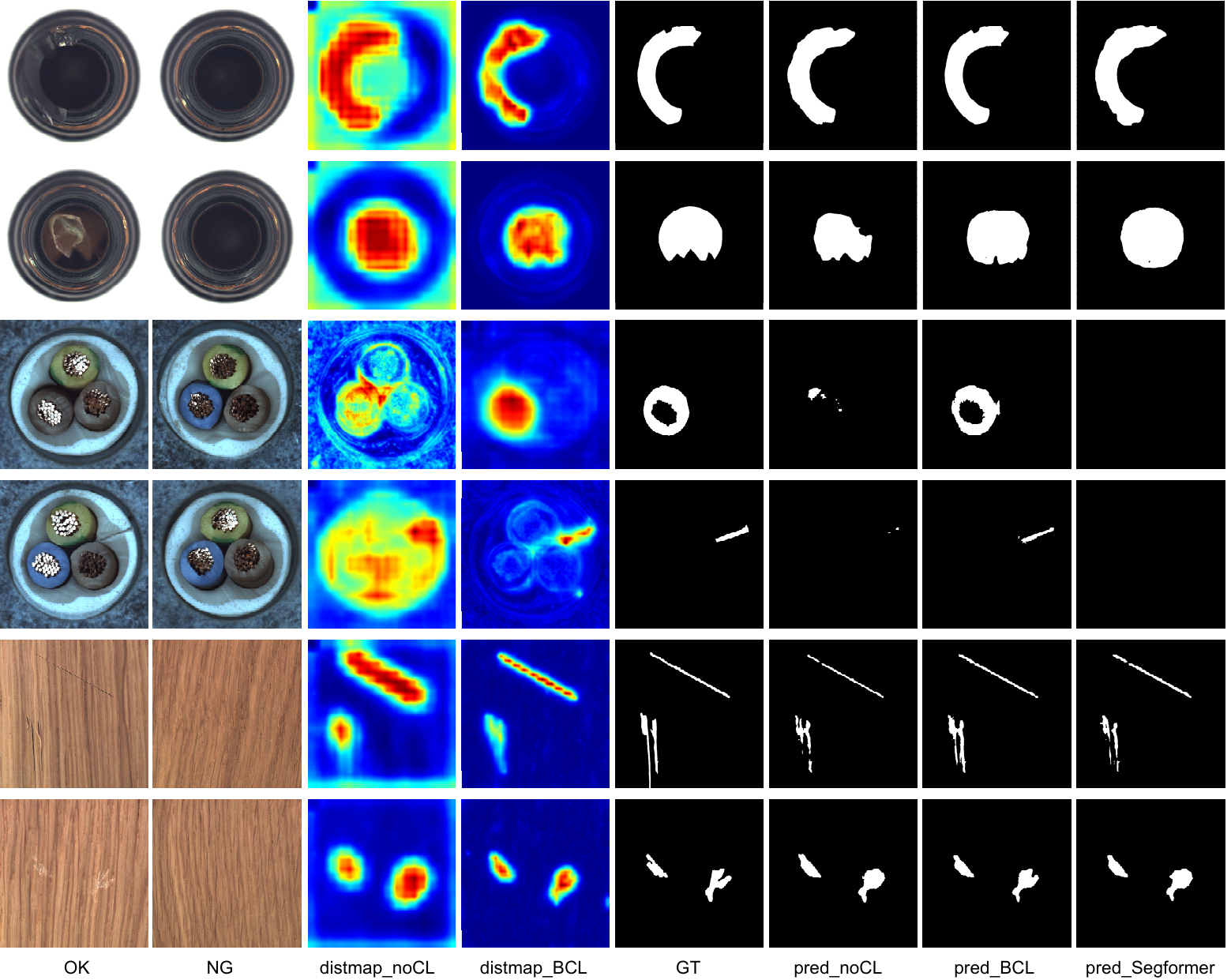}
	\caption{Visual comparison on the MvTec-AD dataset. The results indicate the superior performance of our method with contrastive constraint, especially in scenes with low-constrast and complex background.}\label{fig_vis_mvtec}
\end{figure}

\begin{figure}[!t]%
	\centering
	\includegraphics[width=1.0\textwidth]{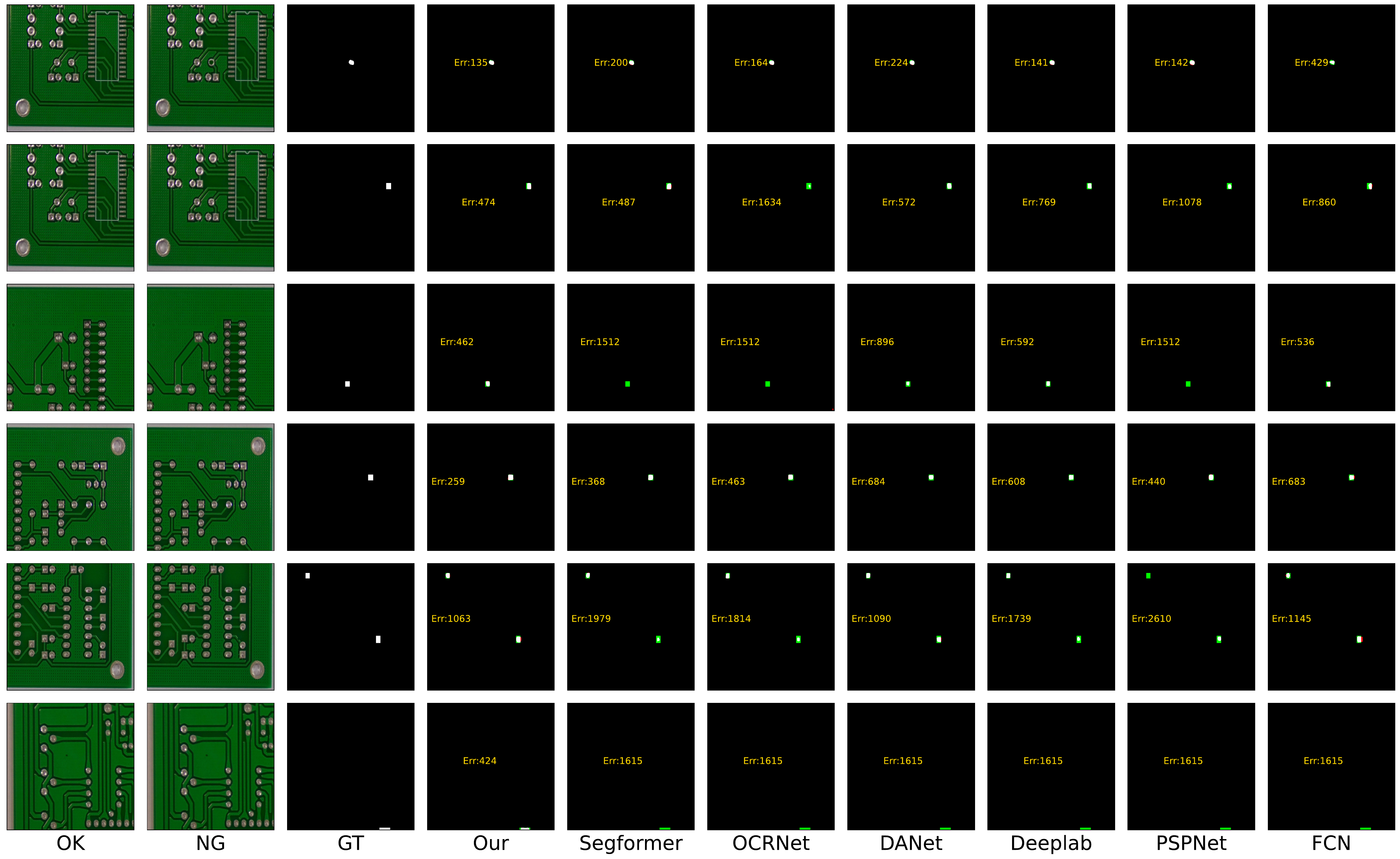}
	\caption{Visual comparison on the PCB dataset. White color represents the defects, while {\color{green}green} color represents missed detections and {\color{red}red} color wrong detections. The errors (Err) in yellow summarise the missed and wrong detections.}\label{fig_vis_pcb}
\end{figure}

\begin{figure}[tbh]%
	\centering
	\includegraphics[width=1.0\textwidth]{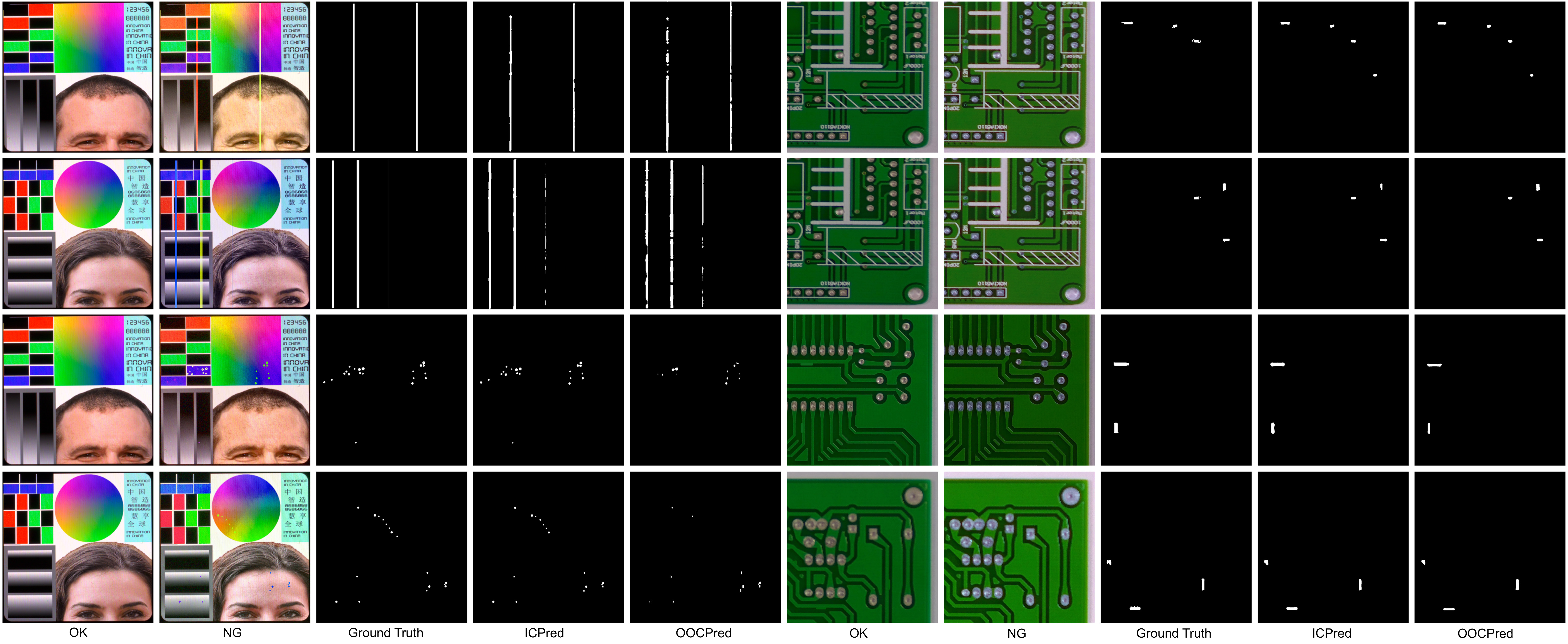}
	\caption{Visualization of the intra-class predictions (ICPred) and out-of-class predictions (OOCPred). }\label{fig7}
\end{figure}

 Figure \ref{fig_vis_mvtec} depicts the results of our method and Segformer in scenes with general industrial products. It is important to note that the high-level semantic defects in rows 3 and 4 cannot be addressed using conventional segmentation methods, as they exhibit normal textures. Figure \ref{fig7} illustrates the intra-class and out-of-class predictions generated by our model. Interestingly, despite the decline in the accuracy of OOC detection, the visual impact is not readily apparent. Indeed, the mIoU values for AL and LA remain impressively robust. Taking the performance of SegFormer on the COCO \cite{lin2014microsoft} and ADE20K \cite{zhou2017scene} datasets as benchmarks, the real-time variant of SegFormer (B0) achieves mIoU scores of 35.6\% and 37.4\%, respectively. The non-real-time version (B5) achieves 46.7\% and 51.0\%, respectively. This comparison underscores the acceptable visual results of AL and LA.

\section{Conclusion}\label{sec5}


Recent advancements in computer vision have improved industrial defect detection, but challenges remain in fine-grained defect segmentation due to limited defect data and inconsistent appearances. To address this, a change-based modeling framework was developed to locate pixel-wise multi-class defects, leveraging the assumption that defective images are formed bt defect-free images. 

We conducted an in-depth comparison between our model and the dense SOTA prediction methods using the SynLCD and two public  datasets. Our model surpasses six leading segmentation models in performance while maintaining reasonable computational costs. Remarkably, our model demonstrates superior out-of-class detection capabilities, in contrast to other segmentation models that produce unsatisfactory results. This breakthrough suggests the feasibility of developing a streamlined approach for basic industrial inspections using only defect-free samples and simulated defects. Furthermore, we evaluated our model with a limited number of labeled samples. Our model's superiority is further underscored when compared with five semi-supervised learning techniques.
Our ablation study demonstrated the effectiveness of the BCL approach, which enhances model performance by applying balanced contrastive constraints and using a change-aware decoder for precise defect localization. The change-aware mechanism aids out-of-class defect detection, which endows our model with considerable potential for real-world applications, especially in scenarios where defect appearances are highly variable.


Several avenues for future research could further enhance our model, including: (1) Exploring advanced data augmentation techniques by GAN and diffusion model, to synthetically expand the defect dataset. This may further improve the model's robustness to unseen defect types. (2) Delving deeper into semi-supervised and unsupervised learning methods that could provide a pathway to leverage unlabeled data more effectively.

{\bf Acknowledgement.} This paper is supported by the ”YangFan” major project in Guangdong province of China, No. [2020] 05.





\let\oldbibitem\bibitem
\renewcommand{\bibitem}[1]{\oldbibitem{#1}\setlength{\itemsep}{0pt}}





\end{document}